# An Effective Algorithm for and Phase Transitions of the Directed Hamiltonian Cycle Problem

**Gerold Jäger**                                                                        GEJ@INFORMATIK.UNI-KIEL.DE
*Computer Science Institute,*
*Christian-Albrechts-University of Kiel,*
*D-24118 Kiel, Germany*

**Weixiong Zhang**                                                                       WEIXIONG.ZHANG@WUSTL.EDU
*Department of Computer Science and Engineering,*
*Washington University,*
*St. Louis, Missouri 63130, United States*

## Abstract

The Hamiltonian cycle problem (HCP) is an important combinatorial problem with applications in many areas. It is among the first problems used for studying intrinsic properties, including phase transitions, of combinatorial problems. While thorough theoretical and experimental analyses have been made on the HCP in undirected graphs, a limited amount of work has been done for the HCP in directed graphs (DHCP). The main contribution of this work is an effective algorithm for the DHCP. Our algorithm explores and exploits the close relationship between the DHCP and the Assignment Problem (AP) and utilizes a technique based on Boolean satisfiability (SAT). By combining effective algorithms for the AP and SAT, our algorithm significantly outperforms previous exact DHCP algorithms, including an algorithm based on the award-winning CONCORDE TSP algorithm. The second result of the current study is an experimental analysis of phase transitions of the DHCP, verifying and refining a known phase transition of the DHCP.

## 1. Introduction

An undirected graph $G = (V, E)$ is *Hamiltonian* if it contains a *Hamiltonian cycle* (HC), a cycle that visits each vertex exactly once. Given a graph, the *Hamiltonian cycle problem* (*HCP*) is to find a HC or to prove that no HC exists in the graph. The decision version of the HCP is among the first problems that were proven to be $\mathcal{NP}$-complete (Karp, 1972). HCP is a well-known problem with many applications in different areas, e.g., the Hamiltonian cycle game in game theory (Stojaković & Szabó, 2005), the problem of finding a knight's tour on a chessboard in artificial intelligence (Henderson & Apodaca, 2008), and the DNA Physical Mapping in biology (Grebinski & Kucherov, 1996). Much research has been done on the HCP in undirected graphs. For reviews, see the work of Bondy (1995), Christofides (1975), Chvátal (1985), Gould (1991), Vandegriend (1998), and Gutin and Moscato (2000). In particular, many algorithms have been developed for the HCP (Angluin & Valiant, 1979; Bollobás, Fenner & Frieze, 1987; Frieze, 1988a; Pósa, 1976; Vandegriend, 1998), as reviewed in the Stony Brook Algorithm Repository (Skiena, 2008). One effective algorithm for the HCP is based on the related Traveling Salesman Problem (TSP) in an undirected weighted graph, which is the problem of finding a HC with minimum total weight.

The HCP is also a canonical problem for understanding intrinsic properties of combinatorial problems. One such problem property is the so called *phase transition*. Consider an undirected graph $G_{n,m}$ with $m$ edges randomly chosen from all possible $n(n-1)/2$ edges over $n$ vertices. It is expected that when keeping the size $n$, i.e., the number of vertices, a constant while increasing the number of edges $m$, the probability that a random graph $G_{n,m}$ is Hamiltonian increases from 0 to 1. Surprisingly, the probability of being Hamiltonian for $G_{n,m}$ exhibits a sharp, dramatic transition from 0 to 1, and the transition occurs approximately when $m = \lceil c \cdot n \cdot (\log n + \log \log n)/2 \rceil$ (Bollobás, 1985; Cheeseman, Kanefsky & Taylor, 1991; Komlós & Szemerédi, 1983). Furthermore, it was experimentally shown that when the constant $c$ is between 1.08 and 1.10, the probability that $G_{n,m}$ is Hamiltonian is 1/2 (Vandegriend & Culberson, 1998). Phase transitions in the HCP have also been studied under other different control parameters, for example, the so called *general constrainedness parameter* (Frank, Gent & Walsh, 1998). The phase transition result of the HCP has motivated a substantial amount of research on phase transitions of other combinatorial problems, particularly the TSP (Zhang & Korf, 1996) and Boolean satisfiability (Monasson, Zecchina, Kirkpatrick & Selman, 1999).

In this study we consider the HCP in directed graphs, which we call *directed HCP*, or *DHCP* for short. In addition to the known applications of the HCP mentioned above, an interesting application of the DHCP is that DHCP heuristics can be used to solve the Bottleneck TSP (Kabadi & Punnen, 2002). In contrast to the extensive amount of work on the HCP for undirected graphs, the research on the DHCP is rather limited (Angluin & Valiant, 1979; Bang-Jensen & Gutin, 2008; Kelly, 2007). The first exact algorithm for the DHCP was developed by Martello (1983). This algorithm outputs a fixed number $h$ of HCs or reports that it cannot find $h$ HCs in a given directed graph. By setting $h = 1$, this gives rise to an algorithm for the DHCP. In recent years, algorithms based on SAT encoding have been introduced to this problem, e.g., the absolute encoding (Hoos, 1999) and the relative encoding (Prestwich, 2003; see also Velev & Gao, 2009). Furthermore, a probabilistic heuristic for DHCP of complexity $\mathcal{O}(n^{1.5})$ was proposed (Frieze, 1988b). It can be shown that for the random class $G_{n,m}$ the probability, that for a given instance a HC is found by this algorithm and therefore exists, changes from 0 to 1, when $n$ grows to infinity and $m = n \log n + cn$, where $c$ is a constant. For the DHCP, a phase transition result similar to that of the HCP has been obtained as well, namely the phase transition occurs at $m = \lceil c \cdot n \cdot (\log n + \log \log n) \rceil$ (McDiarmid, 1980), where the constant $c$ was expected to be close to 1.

Note that the research on the TSP has also alluded to a DHCP algorithm. Using the technique of 2-point reduction, the asymmetric TSP (ATSP) – where the distance from city $i$ to city $j$ may not be necessarily equal to that from $j$ to $i$ – can be converted to the symmetric TSP, with the number of vertices being doubled (Jonker & Volgenant, 1983). Using this transformation, we can determine whether a directed graph is Hamiltonian by solving the symmetric TSP using the renowned CONCORDE algorithm (Applegate, Bixby, Chávatal & Cook, 2005, 2006). CONCORDE has solved many large benchmark instances (Cook, 2010), including a TSP instance with 85,900 cities (Applegate et al., 2009), which up to date is the largest solved practical TSP instance.

The main contribution of this paper is an effective exact algorithm for the DHCP. In our algorithm, we utilize methods for two well-known combinatorial problems, i.e., the Assign-ment Problem (AP) and Boolean satisfiability (SAT); we therefore denote our algorithm by AP-SAT. Using random graphs and many real world instances, we experimentally compare the AP-SAT algorithm with the DHCP algorithm of Martello (1983), the TSP based approach that takes advantage of the TSP solver CONCORDE (Applegate et al., 2005, 2006) and the above-mentioned SAT encodings for the DHCP (Hoos, 1999; Prestwich, 2003). The results show that the AP-SAT algorithm significantly outperforms these algorithms.

The second contribution is an experimental study and refinement of the known phase transition result on the existence of a HC in a random directed graph (McDiarmid, 1980), as similarly done for the HCP (Vandegriend & Culberson, 1998).

## 2. The Algorithm

Consider a directed unweighted graph $G = (V, E)$ with nodes $V$ and edges $E$. For our purpose of solving the DHCP, we consider the problem of determining whether or not there exists a collection of cycles, which may not be necessarily complete cycles, visiting each vertex exactly once. We call this problem *directed Assignment Problem* or *DAP* for short. Our algorithm explores and exploits the intrinsic relationship between the DHCP and the DAP. More precisely, the AP-SAT algorithm searches for a HC in the space of DAP solutions. It first solves the DAP. If the DAP solution forms a HC, or no DAP solution exists, the algorithm terminates. If the DAP solver returns a solution that is not a HC, the algorithm then tries to patch the subcycles in the solution into a HC using the well-known Karp-Steele patching method (Karp & Steele, 1985). If no HC is found either, these DAP and patching steps are iterated, with the only difference that another DAP solution might be found. For most cases that we considered in this study, the algorithm can find a HC or determine that no solution exists after these two steps. If the algorithm fails to solve the problem after these iterative steps, it then attempts to enumerate the DAP solutions by formulating the DAP as a Boolean satisfiability problem and repeatedly solving the problem using a SAT solver and adding constraints to eliminate the DAP solutions that have been encountered. We discuss the details of these steps in the rest of the section.

### 2.1 Solving the Assignment Problem

Given $n$ vertices and a matrix $C = (c_{ij})_{1 \leq i,j \leq n} \in \mathbb{R}^{n,n}$ of the costs between pairs of vertices, the Assignment Problem (AP) is to find a vertex permutation $\pi^*$ such that $\pi^* = \arg\min \left\{ \sum_{i=1}^n c_{i,\pi(i)} : \pi \in \Pi_n \right\}$, where $\Pi_n$ is the set of all permutations of $\{1, \ldots, n\}$. Note that an AP solution can be viewed as a collection of cycles visiting each vertex exactly once.

Many algorithms have been developed for the AP (Bertsekas, 1981; Goldberg & Kennedy, 1995; Jonker & Volgenant, 1987). (For an experimental comparison of AP algorithms see Dell'Amico & Toth, 2000.) The most efficient one is the Hungarian algorithm, which is based on König-Egervary's theorem and has a complexity of $\mathcal{O}(n^3)$. In the AP-SAT algorithm we use the implementation of the Hungarian algorithm by Jonker and Volgenant (1987, 2004).

For an unweighted directed graph $G = (V, E)$, DAP can be solved by applying an AP algorithm to the AP instance defined by the matrix $C = (c_{ij})_{1 \leq i,j \leq n}$ with

$$c_{ij} = \begin{cases} 0, & \text{if } (i,j) \in E,\ i \neq j \\ 1, & \text{if } (i,j) \notin E,\ i \neq j \\ 1, & \text{if } i = j \end{cases}$$

where we map the costs of arcs in $G$ to 0 and the costs of the remaining arcs to 1. If the AP algorithm returns a solution with cost 0, there is a DAP solution in $G$, since every arc taken in the AP solution is an arc in $G$. On the other hand, if it returns a solution of cost greater than 0, there is no DAP solution in $G$ because at least one arc in the solution does not belong to $G$.

The first step of the AP-SAT algorithm is this DAP algorithm. Then a HC of $G$, if one exists, is a solution to the DAP. We have to distinguish three cases at the end of the first step:

- If the cost of the AP solution is greater than 0, $G$ does not have a HC, and the DHCP instance is solved with no solution.

- If the AP solution has cost 0 and the solution consists of one cycle, we have found a HC – and the DHCP instance is also solved.

- If the AP solution has cost 0 and the AP solution has more than one cycle, we cannot determine, based on the AP solution, whether or not $G$ is Hamiltonian. We then continue to the next steps of the AP-SAT algorithm.

### 2.2 Karp-Steele Patching

If the DAP solution does not provide a definitive answer to the problem, i.e., the case where the AP solution cost is 0 and the AP solution contains more than one cycle, we continue to search for a HC in $G$. We first patch the subcycles in an attempt to form a HC, and we use *Karp-Steele patching* (*KSP*) for this purpose, which is an effective ATSP heuristic (Glover, Gutin, Yeo & Zverovich, 2001; Goldengorin, Jäger & Molitor, 2006; Karp & Steele, 1985). The operation of patching two cycles $C_1$ and $C_2$ in an AP solution is defined as follows: two fixed arcs $(v_1, w_1) \in C_1$ and $(v_2, w_2) \in C_2$ are first deleted and two arcs $(v_1, w_2)$ and $(v_2, w_1)$ joining the two cycles are added. The cost of patching $C_1$ and $C_2$ using $(v_1, w_2)$ and $(v_2, w_1)$ is equal to

$$\delta(C_1, C_2) = c(v_1, w_2) + c(v_2, w_1) - (c(v_1, w_1) + c(v_2, w_2))$$

i.e., $\delta(C_1, C_2)$ is the difference between the total cost of the inserted arcs and the total cost of the deleted arcs. In each step we choose to patch the two cycles that have the largest number of vertices. For these two cycles, the two arcs are chosen in such a way that the patching cost is the minimum among all possible arc pairs. If we have $k \geq 2$ cycles, we repeat this patching step $k - 1$ times to form one cycle at the end. We apply KSP to the AP instance defined in Section 2.1. If the patching procedure provides a HC, the AP-SAT algorithm can be terminated. Otherwise, we continue to the next step.

### 2.3 Solving Variant APs

DAP may have multiple solutions, and some of the DAP solutions may be HCs. We can increase the chance of finding a HC if we apply the AP step multiple times, since the computational cost of the AP and the KSP algorithms is low. The key is to avoid finding the same DAP solution again. To accomplish this, we slightly alter some of the arc costs of the corresponding AP instance so as to find the other DAP solutions, enhanced by the KSP if needed, to increase the possibility of finding a HC. In other words, we add a "perturbation" component to create multiple variant AP instances to boost the overall chance of finding a HC. Note that in the worst case when the DHCP instance contains no HC, this procedure will not be productive.

The main idea to create a variant AP instance is to reduce the chance that the subcycles in the current AP solution can be chosen in the subsequent rounds of solving the APs. This is done by "perturbing" the costs of some of the arcs in $G$ as follows. For each arc in the current DAP solution we increase its cost by one. To create an AP instance different from that in Section 2.1, we generalize the AP instance as follows. Let $c_{i,j}$ be the cost of the arc $(i, j) \in E$, and let

$$M \quad := \quad n \cdot \max \left\{ c_{i,j} \,|\, (i,j) \in E \right\} + 1$$

i.e., $M$ is greater than $n$ times the largest cost of an arc in $G$. We then set the costs of the edges not in $E$ to $M$. The AP instance of Section 2.1 is a special case of this AP instance, where the costs $c_{i,j}$ for all arcs $(i, j) \in E$ are 0. It is critical to notice that all DAP solutions, including a HC, must have costs less than $M$. As before, if the solution contains a HC, the algorithm terminates; otherwise, the subcycles are patched using the KSP to possibly find a HC. We repeat this step multiple times so that an arc, which has appeared in many previous DAP solutions, will be very unlikely to appear in the next DAP solution, and an arc, which has never occurred in any previous DAP solution, will be more likely to appear in the next DAP solution.

Let $r$ be the maximal number of AP/KSP calls, i.e., the number of variant AP instances solved. We observed in our experiments that $r = n$ (see step 3 of the pseudo code of the appendix) is a good choice. This will be discussed in detail in Section 3.1.

### 2.4 Implicitly Enumerating all DAP Solutions Using SAT

All the AP and patching based steps discussed above may still miss a solution to a DHCP instance. We now consider how to implicitly enumerate all DAP solutions for finding a solution to the DHCP, if it exists. The idea is to systematically rule out all the DAP solutions that have been discovered so far during the search. To this end, we first formulate a DAP as a Boolean satisfiability (SAT) problem (Dechter, 2003) and forbid a DAP solution by adding new constraints to the SAT model. This elementary technique of adding new constraints with the purpose of enumerating all SAT solutions can also be applied to a general SAT problem (e.g., see Jin, Han & Somenzi, 2005). Notice that this cannot be easily done under the AP framework because such constraints cannot be properly added to the AP. Moreover, we can take advantage of the research effort that has been devoted to SAT, in particular, we can use an effective SAT solver called MINISAT (Eén & Sörensson, 2003, 2010).

In the conjunctive normal form (CNF), a SAT instance over a set of Boolean variables is a conjunction of clauses, each of which is a disjunction of literals which are Boolean variables or their negations. A clause is satisfied if one of its literals is TRUE, and the instance is satisfied if all its clauses are satisfied. The SAT problem is to find a truth assignment of the variables to satisfy all clauses if they are satisfiable, or to determine no such assignment exists. SAT was the first problem shown to be $\mathcal{NP}$-complete (Cook, 1971; Garey & Johnson, 1979; Karp, 1972).

We now formulate the DAP in SAT. A solution to a DAP must obey the following restrictions:

- For each vertex $i$, $i = 1, \ldots, n$, exactly one arc $(i, j)$, $i \neq j$, exists in the DAP solution.

- For each vertex $i$, $i = 1, \ldots, n$, exactly one arc $(j, i)$, $j \neq i$, exists in the DAP solution.

We first introduce an integer decision variable $x_{i,j}$ to the arc $(i, j) \in E$ where $x_{i,j} = 1$ holds if and only if the arc $(i, j)$ appears in the DAP solution. We represent the above constraints in the following integer linear program (ILP).

$$\begin{cases} \sum_{j=1, (i,j) \in E}^{n} x_{i,j} = 1 \text{ for } i = 1, \ldots, n \\ \sum_{i=1, (i,j) \in E}^{n} x_{i,j} = 1 \text{ for } j = 1, \ldots, n \end{cases} \quad (1)$$

where $x_{i,j} \in \{0, 1\}$ for $(i, j) \in E$. We thus have a total of $2n$ constraints. Note that we only have to use $m$ variables, one variable for each arc in the graph, which can be substantially smaller than $n^2$ variables for sparse graphs. We represent the integer linear program (1) by a SAT model similar to the work of Lynce and Marques-Silva (2006), where we replace integer variables $x_{i,j}$ with Boolean variables $y_{i,j}$. To enforce the $2n$ restrictions in the SAT formulation, we need to introduce constraints in clauses. One restriction in (1) means that exactly one of the up to $n$ involved Boolean variables for a vertex can be set to TRUE and the rest must be FALSE. To represent this, we introduce at most $2n^2$ auxiliary variables $z_1, z_2, \ldots, z_{2n^2}$, with up to $n$ $z$'s for one restriction. Without loss of generality, consider the first restriction, which has $z_1, z_2, \ldots, z_n$ associated. We use $z_k$ to represent that at least one of $y_{1,1}, y_{1,2}, \ldots, y_{1,k}$ is TRUE. Precisely, the $z$ variables are defined as follows.

- $z_1 = y_{1,1}$ or equivalently $(\neg y_{1,1} \vee z_1) \wedge (y_{1,1} \vee \neg z_1)$.

- $z_k = y_{1,k} \vee z_{k-1}$ or equivalently $(z_k \vee \neg y_{1,k}) \wedge (z_k \vee \neg z_{k-1}) \wedge (\neg z_k \vee y_{1,k} \vee z_{k-1})$ for $k = 2, 3, \ldots, n$.

In addition, we need to enforce that only one $y_{1,i}$ for $i = 1, 2, \ldots, n$ can be TRUE. This means that if $y_{1,k}$ is TRUE, none of the $y_{1,i}$ for $i < k$ can be TRUE. This can be formulated as

- $\neg z_{k-1} \vee \neg y_{1,k}$ for $k = 2, 3, \ldots, n$.

Finally, $z_n$ must be TRUE. The other restrictions in (1) are represented similarly.

The SAT based representation allows us to exclude a non-Hamiltonian DAP solution previously found in the search. This can be done by introducing new clauses to explicitly forbidding all subcycles of this solution. Let such a subcycle be $(v_1, v_2, \ldots, v_k, v_1)$. Then we add the clause

$$\neg y_{v_1,v_2} \vee \ldots \vee \neg y_{v_{k-1},v_k} \vee \neg y_{v_k,v_1}$$

to the current SAT instance. As a result, the updated SAT instance is not satisfiable, meaning that the corresponding DHCP instance does not contain a HC, or gives rise to a new DAP solution, as it does not allow the previous DAP solution.

In summary, after the AP- and patching-related steps failed to find a solution, the AP-SAT algorithm transforms the problem instance into a SAT instance. Then it collects all previous DAP solutions, each of which includes at least two subcycles, and excludes these subcycles for each of the DAP solutions by adding new clauses as described above. Then the resulting SAT model is solved. If the SAT model is not satisfiable, then the DHCP algorithm terminates with the result of the problem instance being not Hamiltonian. If the SAT model is satisfiable and the solution has only one cycle, the algorithm stops with a HC. If the SAT model is satisfiable, but the solution has more than one subcycle, new clauses are introduced to the SAT model to rule out this solution, and the algorithm repeats to solve the revised formula. Since there is a finite number of DAP solutions, the algorithm terminates. In the worst case if the DAP solutions contain no HC, the SAT part of the algorithm will enumerate all these DAP solutions. For an overview, we outline the main steps of the AP-SAT algorithm in a pseudo code in the appendix.

## 2.5 Some General Remarks

Before we present our experimental results, we like to comment on the method we proposed to help appreciate its features.

1. The AP-SAT algorithm consists of three main components, namely the AP step, the KSP step and the SAT step. It might be interesting to know which of these components is the most important one. For this, we have to distinguish between completeness and efficacy of the algorithm. The only necessary step for the completeness is the SAT step of Section 2.4. This step without all previous steps leads also to a correct DHCP algorithm. On the other hand, the AP-SAT algorithm is more effective if the AP and the KSP steps are called often and the SAT step is not called or called only a few times. For example, if for an instance no DAP solution exists or an existing HC is found by the previous steps, the SAT part will not be invoked at all. Indeed, our experiments showed that the SAT step is not invoked for most of the test instances. Regarding the relative time needed by the AP and the KSP steps, we have to consider the density of problem instances. For an instance with a small number of arcs, in most cases there is not only no HC solution, but also no DAP solution. In this case the algorithm terminates after the first AP step and does not need to make any KSP call. On the other hand, an instance with a large number of arcs should require many AP steps, as many DAP solutions may exist which are not HCs, and thus a HC solution may have to be found by KSP. This expected behavior could be validated by experiments: the time for the KSP steps is smaller for instances with a small number of arcs, but is larger for instances with a large number of arcs (see Figure 4).

2. The AP-SAT algorithm is also able to solve HCP as a special case of DHCP, but it is less effective for this case. The reason is that for a symmetric case, an arc and its reverse arc are often present in a DAP solution, resulting in many small cycles of two vertices in the solution. Thus in general we have to enumerate a large number of DAP solutions. In the worst case when no HC exists, all these DAP solutions have to be enumerated, giving rise to a long running time.

3. We can easily revise the AP-SAT algorithm to identify all HCs in a directed graph. Finding all solutions can be desirable for many applications, e.g., the problem of finding all knight's tour on a chessboard (Henderson & Apodaca, 2008; Kyek, Parberry & Wegener, 1997). For algorithms for this problem, see the already mentioned algorithm of Martello (1983) and the algorithm of Frieze and Suen (1992). The revision works as follows. If no HC exists, the algorithm remains the same. Consider now the case that at least one HC exists. If the first HC has been found, the original AP-SAT algorithm terminates in this case. The revised algorithm at this stage saves the first HC, and then continues to search for the next HC. In the pseudo code of the appendix, we only need to replace "**STOP** with" by "**SAVE**" in rows 8, 11, and 23. Note that for the revised algorithm, the SAT part is always invoked if at least one HC exists. Furthermore – like the original AP-SAT algorithm – this revised algorithm works also for the symmetric case, but is less effective.

4. The AP-SAT algorithm used a restart scheme, i.e., it repeatedly solved a series of AP instances, which were derived by modifying costs of the arcs appeared in the previous AP solution. Although the restart scheme and the random restart scheme, which was developed for constraint problems in artificial intelligence (Gomes, Selman & Kautz, 1998), follow the same design principle of trying to avoid to encounter the same solutions again in subsequent runs, these two schemes are fundamentally different. As its name indicated, the random restart scheme depends on random choices made for variable and value selections in the process of search for a variable assignment for a constraint problem. In contrast, our restart scheme is not random; the arcs in the current AP solution will receive higher costs so that the subcycles in the current AP solution will less likely be chosen again. In other words, the restart scheme we used is somewhat deterministic and depends on solution structures of the problem.

5. The method we used to exclude the subcycles in the solution to the current DAP instance from the subsequent SAT solving process follows in principle the popular idea of adding "no-good" constraints to a constraint satisfaction problem (Frost & Dechter, 1994; Richards & Richards, 2000; Zhang, Madigan, Moskewicz & Malik, 2001). Specifically, these subcycles are forbidden by introducing additional constraints.

## 3. Experimental Results

We have implemented the AP-SAT algorithm, the DHCP algorithm of Martello (1983), the DHCP algorithms based on the absolute SAT encoding (Hoos, 1999) and the relative SAT encoding (Prestwich, 2003) in C++ and compared them to an algorithm based on the award-winning CONCORDE TSP program (Applegate et al., 2005, 2006). For the al-gorithm of Martello we have implemented a version which terminates whenever a HC, if one exists, is found. For the SAT based algorithms we used the AP solver of Jonker and Volgenant (1987, 2004) and the MiniSat SAT solver of Eén and Sörensson (2003, 2010). To apply Concorde, a DHCP instance was first transformed to an asymmetric TSP instance by the transformation in Section 2.1 and then to a symmetric TSP instance by the 2-point reduction method (Jonker & Volgenant, 1983). In our implementation, the 2-point reduction works as follows for a graph $G = (V, E)$ with $V = \{v_1, v_2, \ldots, v_n\}$.

1. Make a copy of the vertices $v_1, v_2, \ldots, v_n$, and create the vertex set $V' := \{v'_1, v'_2, \ldots, v'_n\}$.

2. Define a new complete graph $G'$ on the vertex set $V \cup V'$ with (symmetric) cost function $c' : V \cup V' \to \{0; 1; 2\}$ by

$$
\begin{aligned}
c'(v_i, v'_j) &:= \begin{cases} 0 & \text{for } 1 \leq i = j \leq n \\ 1 & \text{for } 1 \leq i \neq j \leq n,\ (v_i, v_j) \in E \\ 2 & \text{for } 1 \leq i \neq j \leq n,\ (v_i, v_j) \notin E \end{cases} \\
c'(v_i, v_j) &:= 2 \quad \text{for } 1 \leq i \neq j \leq n \\
c'(v'_i, v'_j) &:= 2 \quad \text{for } 1 \leq i \neq j \leq n
\end{aligned}
$$

Then a directed HC exists on $G$ if and only if a TSP tour of cost $n$ exists on $G'$. Note that – in contrast to the general version of the 2-point reduction – no value of $-\infty$ is required here. We also tried the 3-point reduction method, which is in principle similar to the 2-point reduction, but uses two (instead of one) copies of the vertex set and uses only cost values from $\{0; 1\}$. For the details of the 3-point reduction, see the work of Karp (1972). Our experimental results, which are not included here, showed that the 3-point reduction runs slower on average than the 2-point reduction. Therefore, in the rest of the comparison, we only consider the 2-point reduction.

After the 2-point reduction, Concorde started with the worst possible solution value as the initial upper bound and was terminated as soon as its lower bound indicates a HC is impossible.

In addition to this comparison, we also experimentally analyzed the AP-SAT algorithm including its asymptotic behavior, and applied it to study phase transitions of the DHCP. All our experiments were carried out on a PC with an Athlon 1900MP CPU with 2 GB of memory.

### 3.1 Comparison of DHCP Algorithms

In our experiments we first tested random asymmetric instances $G_{n,m}$ and parameters $n = 100, 200, 400, 800, 1600$ and $m = \lceil c \cdot n \cdot (\log n + \log \log n) \rceil$ with $c = 0.5, 0.6, \ldots, 1.90, 2.00$. For each $n$ and each $c$ we generated 50 random instances and measured the CPU time for these instances. Furthermore, we tested real-world and random instances from the Dimacs challenge (Johnson et al., 2002, 2008) and non-random instances (Reinelt, 1991, 2008). Whereas Tsplib contains 26 single asymmetric TSP instances with sizes from 17 to 443, the Dimacs challenge contains 10 asymmetric problem generators called *amat, coin, crane, disk, rect, rtilt, shop, stilt, super*, and *tmat*. Using each of these generators we generated 24 instances, 10 with 100 vertices, 10 with 316 vertices, 3 with 1000 vertices, and 1 with 3162 vertices, leading to 240 instances (for each of the 10 problem generators 24 instances) overall. To transform asymmetric TSP instances back to DHCP instances, it seems to be reasonable to only keep the arcs of small weights while ignoring the ones with large weights. In other words, to generate a DHCP instance we chose the $m$ smallest arcs in the corresponding asymmetric TSP instance. It is interesting to note that the most difficult problem instances for most problems in Tsplib and Dimacs appear when the degree parameter $c$ is around 2, which is the value we used in our experiments. In contrast, the most difficult instances of random graphs occur when the degree parameter $c$ is 0.9 (see Section 3.3).

To investigate the variation of running time, we present one subfigure for each problem class, i.e., for the 5 random classes with sizes 100, 200, 400, 800, 1600, and for the 10 Dimacs classes *amat, coin, crane, disk, rect, rtilt, shop, stilt, super*, and *tmat*. The $y$-axis gives the average times plus their 95% confidence intervals, where all values are in seconds. For the random classes the $x$-axis describes the degree parameter $c$, and for the Dimacs classes it describes the size $n$. The results for the random instances are summarized in Figure 1 and for the Dimacs instances in Figures 2, 3. As the Tsplib class consists only of 26 single instances with completely different sizes, structures and difficulties, we present these results in Table 1. If an experiment of a single algorithm on a single instance required at least 1 hour or did not terminate due to a high memory requirement, we set the CPU times as "3600 seconds".

Figures 1 – 3 and Table 1 show that the two SAT encodings are not competitive with AP-SAT, Concorde or the Martello algorithm. Furthermore, AP-SAT and Concorde are more stable than the Martello algorithm. Concorde failed to solve 16 Dimacs instances (3 *coin*, 3 *crane*, 4 *rect*, 5 *stilt*, 1 *super* types) within the maximal allowed time of 1 hour, whereas the AP-SAT algorithm failed only on 7 instances. Among these 7 instances on which AP-SAT failed, 6 are *stilt* types, and the remaining instance (super3162) could be solved if we increased the maximal allowed time from 1 hour to 4 hours (see Table 2). The Martello algorithm was unable to solve the instances with 800 or larger size because of its high memory requirement. For the other instances, it failed on 1 random instance of size 400 with degree parameter 0.9, on 51 Dimacs instances (10 coin, 12 crane, 11 disk, 11 rect, 7 stilt types), and 9 Tsplib instances (see Table 1). Nevertheless, the Martello algorithm outperformed Concorde on smaller and easier instances, indicating that the former has a worse asymptotic running time. Overall, we observed that the AP-SAT algorithm is clearly superior to the four other algorithms. Among the 4266 instances (4000 random instances, 240 Dimacs instances and 26 Tsplib instances) tested, only on 13 instances, one of the other four algorithms is faster than AP-SAT. These problem instances include 4 random instances, namely 1 of size 400 with degree parameter 0.9, 3 of size 800 with degree parameters 0.8, 0.9, 0.9, respectively, 8 Dimacs instances, namely coin1000-2, rect316-9, stilt100-1, stilt100-5, stilt100-6, stilt100-7, stilt100-8, stilt316-2, and the Tsplib instance br17 (see Table 1).

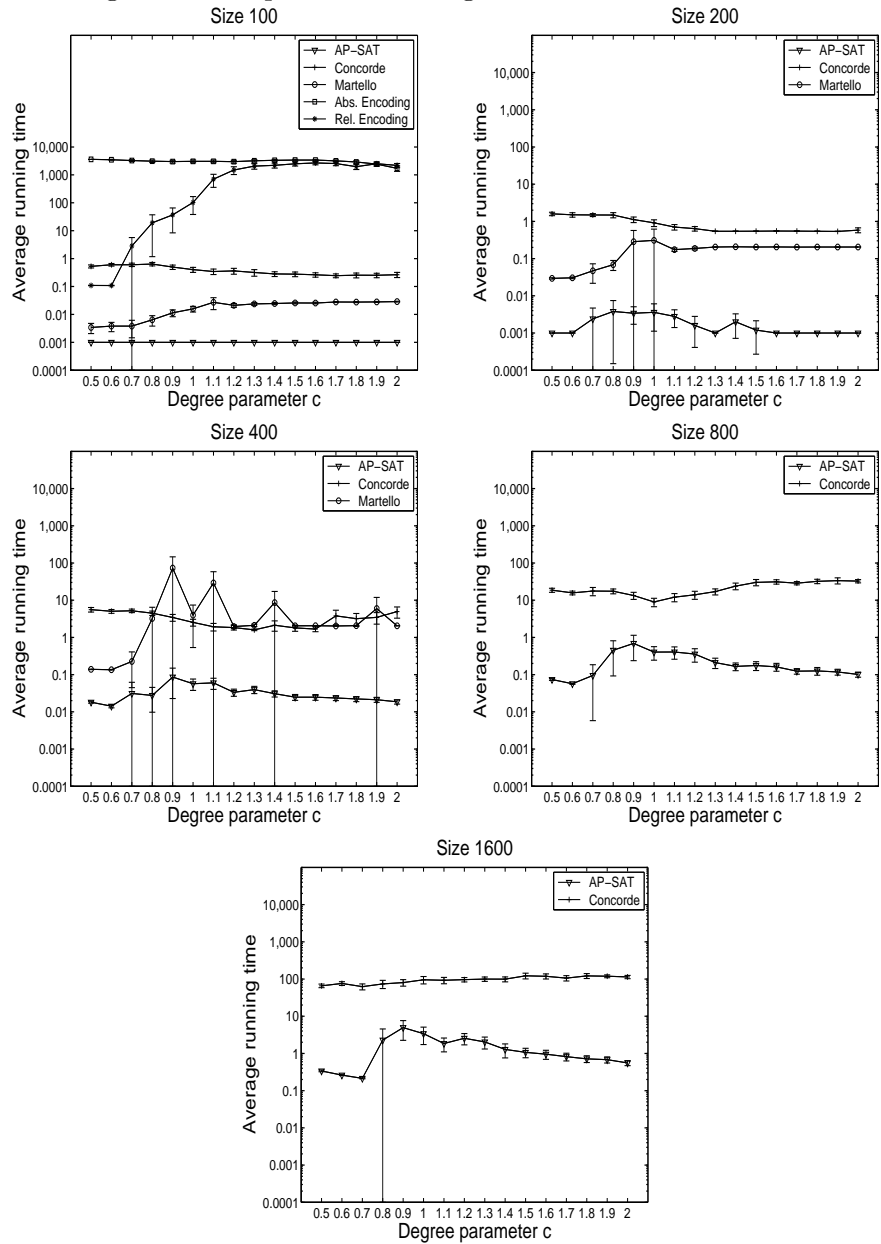

Figure 1: Comparison of all algorithms on random instances.

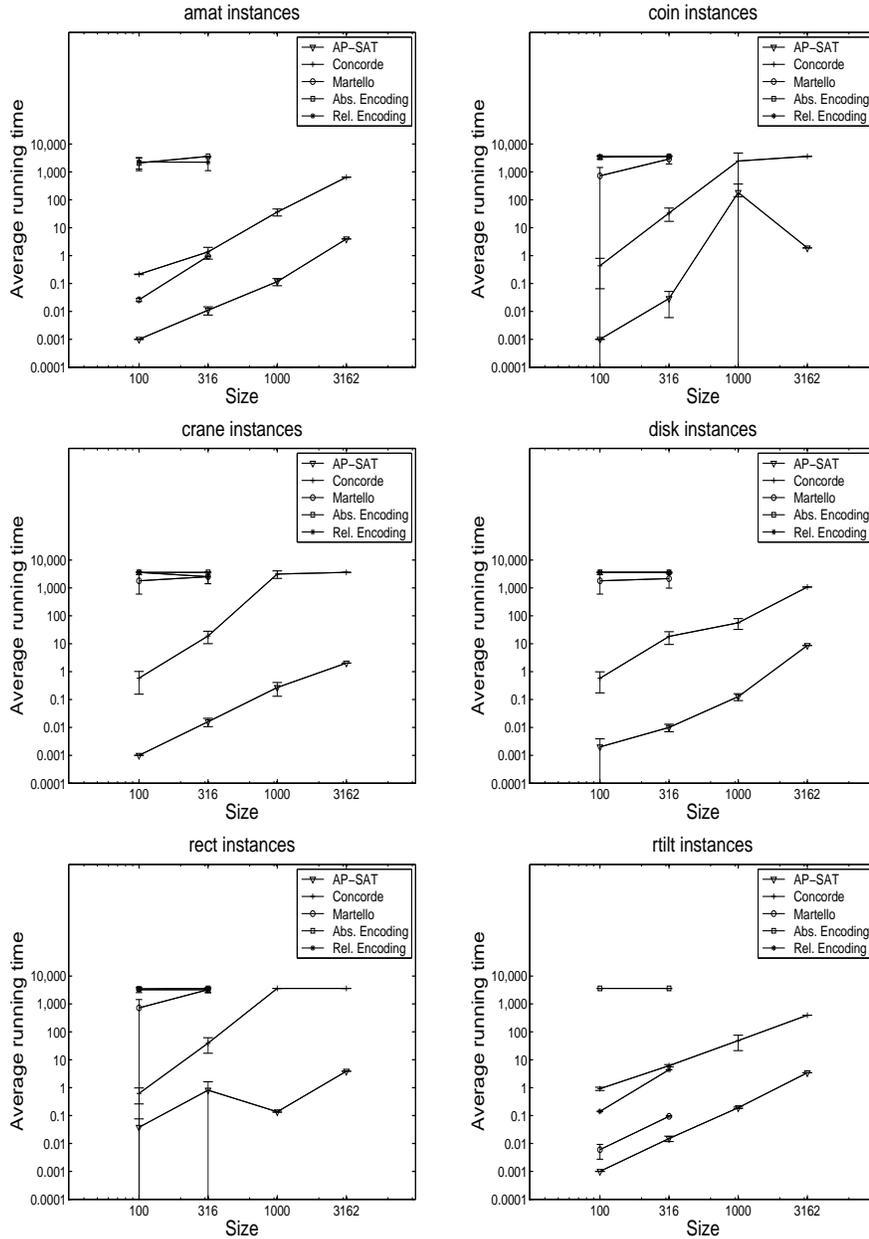

Figure 2: Comparison of all algorithms on Dimacs instances, part 1.

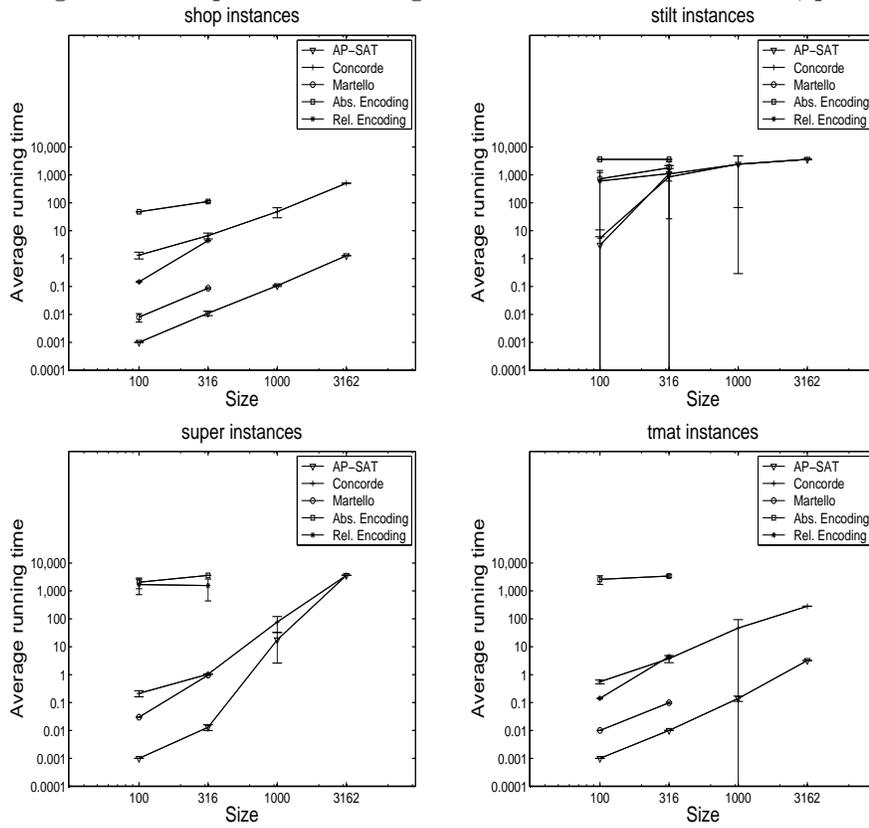

Figure 3: Comparison of all algorithms on Dimacs instances, part 2.

Table 1: Comparison of all algorithms on Tsplib instances.

| Instance (Size) | Running time for algorithm | | | | |
|---|---|---|---|---|---|
| | AP-SAT | Concorde | Martello | Absolute Encoding | Relative Encoding |
| br17 (17) | 4.05 | 0.07 | 66.39 | 0.85 | 0.08 |
| ftv33 (34) | 0 | 0.13 | 0 | 3.29 | 0 |
| ftv35 (36) | 0 | 0.15 | 0 | 5.59 | 0 |
| ftv38 (39) | 0 | 0.23 | 0 | 2.96 | 0.01 |
| p43 (43) | 0 | 0.42 | 0 | 65.58 | 0.01 |
| ftv44 (45) | 0 | 0.19 | 0 | 1.97 | 0.01 |
| ftv47 (48) | 0 | 0.16 | 0 | 5.23 | 0.01 |
| ry48p (48) | 0 | 0.07 | 3600 | 53.96 | 47.91 |
| ft53 (53) | 0 | 1.56 | 0 | 20.31 | 0.02 |
| ftv55 (56) | 0 | 0.09 | 0.01 | 11.13 | 230.04 |
| ftv64 (65) | 0 | 0.23 | 0 | 119.96 | 0.04 |
| ft70 (70) | 0 | 0.87 | 0 | 34.18 | 0.05 |
| ftv70 (71) | 0 | 0.29 | 0 | 1904.56 | 0.06 |
| kro124p (100) | 0 | 3.74 | 0.04 | 1993.33 | 3600 |
| ftv100 (101) | 0 | 0.56 | 3600 | 1024.22 | 3600 |
| ftv110 (111) | 0 | 2.42 | 3600 | 3600 | 3600 |
| ftv120 (121) | 0 | 0.8 | 3600 | 3600 | 3600 |
| ftv130 (131) | 0 | 3.04 | 3600 | 3600 | 3600 |
| ftv140 (141) | 0 | 0.84 | 3600 | 593.65 | 3600 |
| ftv150 (151) | 0 | 1.13 | 3600 | 2676.16 | 3600 |
| ftv160 (161) | 0 | 1.14 | 3600 | 3600 | 3600 |
| ftv170 (171) | 0 | 2.13 | 3600 | 3600 | 3600 |
| rbg323 (323) | 0.01 | 4.81 | 0.09 | 3600 | 5.12 |
| rbg358 (358) | 0.02 | 13.55 | 0.14 | 3600 | 6.98 |
| rbg403 (403) | 0.02 | 4.52 | 0.18 | 3600 | 10 |
| rbg443 (443) | 0.02 | 6.73 | 0.21 | 3600 | 13.24 |

### 3.2 Analysis of AP-SAT

The efficacy of the AP-SAT algorithm may be due to the following reasons. Instances with no HC are most likely to have no DAP solution either, and therefore the algorithm terminates after the first AP call. On the other hand, instances with a HC are likely to have multiple HCs, one of which can be found quickly by the AP or KSP steps. The only difficult case is when there are many DAP solutions, but none or a very few of them are HCs. In this case the AP and KSP steps may fail, and the SAT part will be invoked to find a HC or to disprove the existence of a HC.

In the following we will analyze the instances where AP-SAT fails or requires much time, and analyze the number $r$ of computing variant AP instances (which we had set to the size of the instance $n$; see the end of Section 2.3). Therefore we investigated the three procedures in AP-SAT, namely AP, KSP and SAT. We observed that the SAT part was invoked only on 14 out of all 4266 instances tested. We considered these 14 and other two instances (stilt3162 and super3162), on which AP-SAT did not terminate in 1 hour, to be hard. To further analyze these 16 hard instances we increased the maximal allowed time from 1 hour to 4 hours. In Table 2 we present the running times of AP, KSP and SAT, and the number of calls to the three procedures, where the numbers of AP and KSP calls are given in the same column, as these two numbers are equal or different by only one (see the pseudo code in the appendix). Furthermore, we add two additional pieces of information: whether an instance has a HC or whether this is unknown, and whether AP-SAT terminated on the instance in 4 hours. In Table 2, "Memory" means that this part terminated due to a high memory requirement. Note that the solution status of the instance stilt316-2 ("no HC") was known, since Concorde – in contrast to AP-SAT – was able to solve it.

Table 2 shows that the running time of AP/KSP contributed to the majority of the total running time of AP-SAT only on 4 out of the 16 hard instances, i.e., coin1000-2 and rect316-9, and the two instances stilt3162 and super3162 on which SAT is not invoked at all. On 6 instances, AP-SAT did not terminate. On 5 out of these 6 instances, i.e., stilt316-2, stilt316-4, stilt316-5, stilt1000-1, and stilt1000-2, the SAT part did not terminate in a reasonable amount of time or the algorithm stopped due to a high memory requirement of SAT.

In order to determine $r$, we re-ran all instances in Table 2 with three different values of $r$, i.e., $r = 0$, $r = n/2$, and $r = 2n$. The results (not presented) showed that when AP-SAT was unable to terminate with $r = n$ (i.e., on the 6 instances stilt316-2, stilt316-4, stilt316-5, stilt1000-1, stilt1000-2, and stilt3162), it also failed to stop with other values of $r$. For all remaining 10 instances, increasing $r = n$ to $r = 2n$ did not reduce the running times. This is reasonable for the two instances coin1000-2 and rect316-9 with a large AP/KSP time, as they have no HC. On the other hand, these two instances are the only ones on which AP-SAT ran faster by using smaller values of $r$, namely coin1000-2 by using $r = n/2$ and rect316-9 by using $r = 0$.

We thus conclude that $r$ should not be increased, but rather be decreased. As it is hard to estimate the memory requirements and the time of the SAT part, one alternative for difficult instances would be to start AP-SAT with a smaller parameter $r$ and then to stop the SAT part after some time or after one unsuccessful call. After that the complete AP-SAT algorithm can be restarted with a larger $r$. For most instances, however, the choice of $r$ is not relevant. More difficult problem instances are required to perform a comprehensive analysis of $r$.

Table 2: Comparison of the performance of AP, KSP, and SAT procedures in the AP-SAT algorithm on 16 hard instances.

| Instance | Running time | | | Number of calls | | HC | Termin. |
|---|---|---|---|---|---|---|---|
| | AP | KSP | SAT | AK/KSP | SAT | | |
| br17 | 0 | 0 | 4.1 | 17 | 138 | No | Yes |
| coin1000-2 | 352.73 | 47.54 | 1.82 | 1000 | 1 | No | Yes |
| rect100-2 | 0.07 | 0.01 | 0.27 | 100 | 1 | No | Yes |
| rect316-9 | 3.69 | 0.61 | 0.35 | 316 | 1 | No | Yes |
| stilt100-1 | 0.17 | 0 | 0.07 | 100 | 1 | No | Yes |
| stilt100-5 | 0.2 | 0.01 | 28.84 | 100 | 41 | Yes | Yes |
| stilt100-6 | 0.17 | 0.02 | 0.07 | 100 | 1 | No | Yes |
| stilt100-7 | 0.17 | 0.05 | 0.06 | 100 | 1 | No | Yes |
| stilt100-8 | 0.15 | 0.03 | 0.07 | 100 | 1 | No | Yes |
| stilt316-2 | 20.21 | 1.37 | 14378.42 | 316 | 1 | No | No |
| stilt316-4 | 13.15 | 0.71 | Memory | 316 | 1 | Unknown | No |
| stilt316-5 | 21.14 | 1.63 | Memory | 316 | 1 | Unknown | No |
| stilt1000-1 | 1446.63 | 107.06 | 12846.31 | 1000 | 1 | Unknown | No |
| stilt1000-2 | 1457.76 | 102.21 | 12840.03 | 1000 | 1 | Unknown | No |
| stilt3162 | 13832.40 | 567.60 | 0 | 650 | 0 | Unknown | No |
| super3162 | 13244.88 | 441.46 | 0 | 1032 | 0 | Yes | Yes |

Figure 4: Comparison of the performance of AP and KSP procedures in AP-SAT on random instances of size 1600.

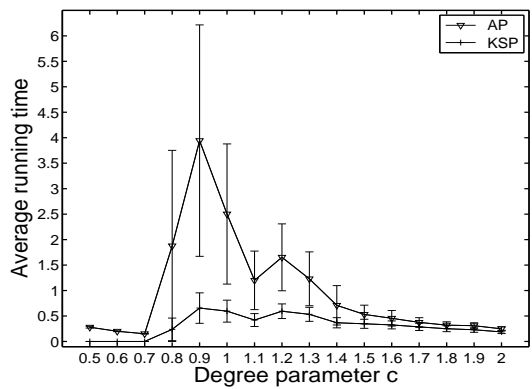

Finally, in Figure 4 we compare the times used by AP and KSP on random instances of size 1600 with degree parameter $c = 0.5, 0.6, \ldots, 1.90, 2.00$.

We observe that AP is more time consuming than KSP. With a smaller $c$ this effect is more obvious because most instances can be solved with a result of "No HC" after the first AP call, and thus the KSP does not need to be invoked at all.

### 3.3 Phase Transitions of the DHCP

For random undirected graphs $G_{n,m}$, where $m$ arcs are randomly chosen from all possible $n(n-1)/2$ arcs over $n$ vertices in the graph, Komlós and Szemerédi (1983) proved a phase transition of $c \cdot \lceil n \cdot (\log n + \log \log n)/2 \rceil$ with $c = 1$ for the HCP. Vandegriend and Culberson (1998) experimentally verified the theoretical result, where the constant $c$ is between 1.08 and 1.10. For the DHCP, where $m$ arcs are randomly chosen from all possible $n(n-1)$ arcs, McDiarmid proved a phase transition of $m = c \cdot \lceil n \cdot (\log n + \log \log n) \rceil$ with $c = 1$ (1980). Our experiments were aimed to verify this result and determine the multiplicative constant $c$. As a directed graph may contain twice as many arcs as the undirected counterpart, we would expect the number of arcs to be doubled as well at the phase transition point. Therefore we tested $m = \lceil c \cdot n \cdot (\log n + \log \log n) \rceil$ with $c = 0.5, 0.6, 0.7, 0.8, 0.81, 0.82, \ldots, 1.19, 1.20,$ 1.30, 1.40, 1.50, 1.60, 1.70, 1.80, 1.90, 2.00, where we expected the phase transition to occur at $c = 1$. We considered problem instances with $n = 128, 256, 512, 1024, 2048, 4096, 8192$ vertices and chose 1000 independently generated random graphs for each $n$ and for each $c$.

The phase transition result is shown in Table 3 and Figure 5, where the first parameter is $c$ and the second parameter the percentage of Hamiltonian graphs among all graphs considered. We observe a phase transition of the DHCP similar to that of the HCP. In particular, it is evident from Figure 5 that the phase transition becomes sharper, i.e., there is a crossover among the phase transition curves, when the problem size increases, which is characteristic for phase transitions in complex systems. This crossover occurs around the degree parameter $c = 0.9$, which is substantially different from the expected value of 1. In short, our observations verified the existence of a phase transition of the DHCP, and the phase transition occurs at $\lceil c \cdot n \cdot (\log n + \log \log n) \rceil$ with approximately $c = 0.9$. Furthermore, for the same constant $c = 0.9$, the probability that $G_{n,m}$ is Hamiltonian is $1/2$. As a comparison, for undirected graphs, a constant between 1.08 and 1.10 was found (Vandegriend & Culberson, 1998).

### 3.4 Asymptotic Behavior of AP-SAT

An interesting characteristic of an algorithm is its asymptotic behavior. To quantify this behavior for the AP-SAT algorithm, we revisited the experiments of Section 3.3, i.e., the experiments that verified the phase transitions of the DHCP. As described earlier, we considered random problem instances with $n = 128, 256, 512, 1024, 2048, 4096, 8192$ vertices and chose 1000 independently generated random graphs for each $n$ and for each $c$. To measure the worst-case asymptotic behavior of AP-SAT, we only measured the CPU times of the algorithm on the most difficult instances, i.e., the instances with degree parameter $c = 0.9$ (see Section 3.3). The results can be found in Figure 6, where the $x$-axis is the problem size and the $y$-axis the average time required. Since both, $x$- and $y$-axis are in logarithmic scale and the log-log curve in Figure 6 is nearly linear, the average running time of AP-SAT can be considered to be polynomial on the number $n$ of vertices in the graph. This is reasonable, as for random instances the SAT part was not called at all (see Section 3.2), and the AP and KSP combined has a complexity not worse than $\mathcal{O}(n^3)$.

Table 3: Phase transition of random instances.

| | Size | | | | | | |
|---|---|---|---|---|---|---|---|
| $c$ | 128 | 256 | 512 | 1024 | 2048 | 4096 | 8192 |
| 0.5 | 0 | 0 | 0 | 0 | 0 | 0 | 0 |
| 0.6 | 0 | 0 | 0 | 0 | 0 | 0 | 0 |
| 0.7 | 5.1 | 3 | 2.6 | 1.3 | 0.5 | 0.1 | 0.2 |
| 0.8 | 23.3 | 21.9 | 21.5 | 18.7 | 16.3 | 14.1 | 12.7 |
| 0.81 | 25.5 | 24.2 | 23.9 | 20.6 | 20.8 | 17.9 | 15.7 |
| 0.82 | 27.8 | 28.4 | 27.4 | 23.1 | 25 | 20 | 18.7 |
| 0.83 | 30.6 | 28.9 | 33 | 25.8 | 27 | 23.3 | 23.9 |
| 0.84 | 32.4 | 31.8 | 34.9 | 32.9 | 28.5 | 29 | 28.9 |
| 0.85 | 33.6 | 34.1 | 35.6 | 30 | 34.3 | 33.4 | 30 |
| 0.86 | 37.6 | 36.5 | 36.9 | 35.4 | 37.4 | 34.7 | 34.1 |
| 0.87 | 39.8 | 39.1 | 38.3 | 40.8 | 41 | 35.9 | 38.7 |
| 0.88 | 44 | 43.5 | 42.4 | 43.8 | 44.7 | 40.2 | 40.6 |
| 0.89 | 46.8 | 47.3 | 47.3 | 45.8 | 47.9 | 47.7 | 44.8 |
| 0.9 | 49 | 52 | 49.8 | 52.5 | 50.1 | 48.6 | 50.5 |
| 0.91 | 52.8 | 53.2 | 52.3 | 54.7 | 50.3 | 52.7 | 54.6 |
| 0.92 | 55.5 | 54.4 | 56.9 | 54.1 | 54.7 | 59.4 | 59.8 |
| 0.93 | 59.2 | 58.4 | 59.5 | 60.8 | 58.4 | 60.7 | 61.5 |
| 0.94 | 60.1 | 60.1 | 61.5 | 63.6 | 61.4 | 65.6 | 65.8 |
| 0.95 | 62.7 | 61.7 | 60.8 | 64.9 | 68.4 | 68.3 | 70.5 |
| 0.96 | 63.6 | 65.3 | 64 | 66.2 | 66.8 | 72.7 | 73.1 |
| 0.97 | 67.2 | 65 | 66.2 | 67.9 | 71.6 | 71 | 71.4 |
| 0.98 | 68.8 | 67.9 | 68.3 | 71.2 | 72.5 | 75.7 | 73.4 |
| 0.99 | 69.4 | 71.8 | 72.1 | 73.8 | 75 | 77 | 76.3 |
| 1 | 71 | 74 | 72.1 | 73.6 | 77.2 | 79.8 | 80 |
| 1.01 | 72.8 | 74.3 | 74.5 | 78.3 | 80.8 | 78.7 | 81.7 |
| 1.02 | 74.2 | 75.4 | 76.3 | 81.1 | 81.4 | 82.4 | 81.9 |
| 1.03 | 75.8 | 75.4 | 79 | 81.9 | 81 | 83.4 | 85.4 |
| 1.04 | 76.9 | 76.6 | 82.1 | 83.2 | 84.2 | 85.3 | 86.5 |
| 1.05 | 77.4 | 78.6 | 84.4 | 85.2 | 86.3 | 88.3 | 88.3 |
| 1.06 | 78.4 | 79.3 | 86 | 85.9 | 85.6 | 85.9 | 90.7 |
| 1.07 | 81.7 | 79.2 | 87.3 | 88.1 | 89.9 | 90 | 92.3 |
| 1.08 | 81.6 | 82.4 | 88.4 | 87.1 | 89.4 | 90.3 | 92 |
| 1.09 | 83.3 | 83.2 | 88.6 | 87.2 | 89.4 | 92.6 | 92.4 |
| 1.1 | 85.1 | 84.4 | 88.8 | 86.8 | 89.5 | 92 | 93.8 |
| 1.11 | 85.4 | 87 | 89.3 | 90.9 | 92 | 93.8 | 93.9 |
| 1.12 | 86.3 | 87.4 | 89.7 | 90.3 | 92.9 | 93.3 | 94 |
| 1.13 | 86.6 | 88.6 | 90.1 | 89.5 | 93 | 93.9 | 94.7 |
| 1.14 | 86.7 | 88.9 | 90.7 | 92.2 | 93.9 | 94.1 | 97.3 |
| 1.15 | 87.3 | 89.2 | 90.9 | 93.2 | 93.8 | 94.2 | 96.4 |
| 1.16 | 88.5 | 89.2 | 92.6 | 93.9 | 95.1 | 95.5 | 97.2 |
| 1.17 | 88.7 | 91.2 | 93.1 | 94.1 | 95.3 | 94.7 | 97.2 |
| 1.18 | 87.8 | 92.6 | 93.6 | 95 0 | 96.1 | 95.8 | 96.4 |
| 1.19 | 88.5 | 92.9 | 93 | 95.5 | 94.8 | 97.3 | 97.2 |
| 1.2 | 89.1 | 93.8 | 93.3 | 96.2 | 96.2 | 97.6 | 97.9 |
| 1.3 | 95.7 | 97.2 | 97.5 | 98.8 | 98.9 | 98.7 | 99.2 |
| 1.4 | 96.4 | 98.8 | 99 | 99.5 | 99.7 | 99.5 | 99.8 |
| 1.5 | 99.1 | 99.1 | 99.9 | 99.9 | 99.8 | 99.9 | 100 |
| 1.6 | 99.7 | 99.6 | 99.8 | 99.9 | 99.9 | 100 | 100 |
| 1.7 | 99.7 | 99.8 | 100 | 100 | 99.9 | 99.9 | 99.9 |
| 1.8 | 99.8 | 99.8 | 99.9 | 99.9 | 100 | 100 | 100 |
| 1.9 | 100 | 100 | 99.9 | 99.9 | 100 | 100 | 100 |
| 2 | 100 | 100 | 100 | 100 | 100 | 100 | 100 |

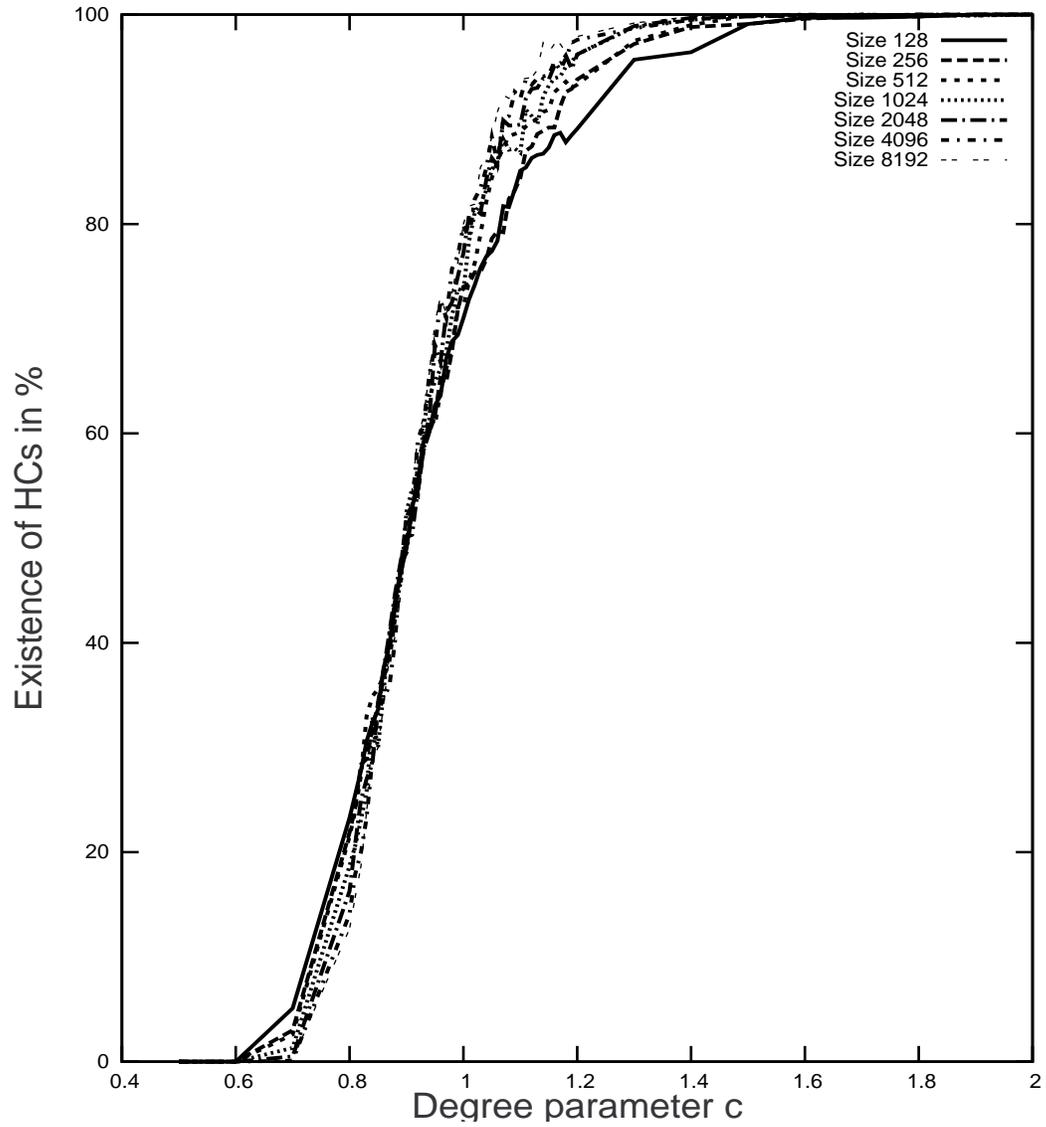

Figure 5: Phase transition of random instances.

Figure 6: Asymptotic behavior of the AP-SAT algorithm.

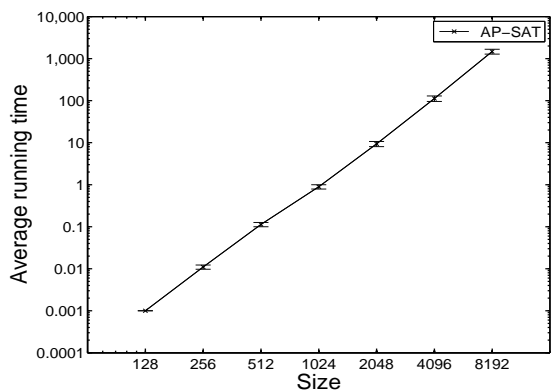

## 4. Summary

The Hamiltonian cycle problem (HCP) is an important, canonical combinatorial problem. Surprisingly, for the HCP in directed graphs, which we called directed HCP or DHCP, no effective exact algorithm has been developed. Our main result of this work is a novel and effective exact algorithm for the DHCP. Our algorithm utilizes an existing algorithm for the assignment problem and an existing method for Boolean satisfiability (SAT). Our work includes a new SAT formulation of the HCP and the AP, which can be potentially extended to other problems such as the TSP. Our experimental results on random and real problem instances showed that our new algorithm is superior to four known algorithms including one algorithm that takes advantage of the award-winning Concorde TSP algorithm. Furthermore, the first phase transition result on combinatorial problems was done on the HCP and later was extended to the DHCP. In this paper we experimentally verified the existence of a phase transition of the DHCP and refined the location where such a phase transition appears using our new exact DHCP algorithm.

## Acknowledgments

We thank David S. Johnson at AT&T Labs - Research and Gregory Gutin at Royal Holloway University of London for many discussions related to this work and their insightful comments on our manuscript. This research was supported in part by NSF grants IIS-0535257 and DBI-0743797 to Weixiong Zhang.

## Appendix A. Pseudo Code of AP-SAT Algorithm

**INPUT** Directed non-complete graph $G = (V, E)$ with $|V| = n$.

1. Define matrix $C$ as in Section 2.1, $M := 1$.
2. Define subcycle collection set $W := \emptyset$.
3. **FOR** $s = 1, \ldots, n$
4.    Solve AP on instance matrix $C$ with solution value $g$, AP solution $(v_1, v_{i_1}), (v_2, v_{i_2}) \ldots, (v_{n-1}, v_{i_{n-1}}), (v_n, v_{i_n})$, number of cycles $k$.
5.    **IF** $g \geq M$
6.       **THEN STOP** with **No HC**.
7.       **ELSE IF** $k = 1$
8.          **THEN STOP** with **HC** being the AP solution.
9.    Apply KSP to the cycles, and receive solution value $h$ and complete cycle $(w_1, w_2, \ldots, w_n, w_1)$.
10.    **IF** $h = 0$
11.       **THEN STOP** with **HC** $(w_1, w_2, \ldots, w_n, w_1)$.
12.    **FOR** $t = 1, \ldots, n$
13.       $c_{v_t, v_{i_t}} = c_{v_t, v_{i_t}} + 1$
14.    $M = n \cdot \max\{c_{i,j} \,|\, (i,j) \in E\} + 1$.
15.    $c_{i,j} = M$ for all $(i,j) \notin E$.
16.    Add each subcycle of AP solution to $W$.
17. Start with the SAT model explained in Section 2.4.
18. For each subcycle $(v_1, v_2, \ldots, v_{k-1}, v_k, v_1)$ of $W$ add the clause $\neg y_{v_1, v_2} \vee \ldots \vee \neg y_{v_{k-1}, v_k} \vee \neg y_{v_k, v_1}$ to the SAT model.
19. Solve the SAT model.
20. **IF** Variable setting exists for the model.
21.    **THEN** Add all $k$ subcycles of the solution of the SAT model to $W$.
22.       **IF** $k = 1$
23.          **THEN STOP** with **HC** being the only subcycle.
24.       **GOTO** 19.
25.    **ELSE STOP** with **No HC**.

**OUTPUT** **HC** of $G$, or proof that **No HC** exists in $G$.